\def\arxivbuild{}
\documentclass{article}

\usepackage{microtype}
\usepackage{graphicx}
\usepackage{subcaption}
\usepackage{booktabs}
\usepackage{hyperref}

\ifdefined\arxivbuild
\usepackage[preprint]{icml2026}
\else
\usepackage{icml2026}
\fi

\usepackage{amsmath}
\usepackage{amssymb}
\usepackage{mathtools}
\usepackage{amsthm}
\usepackage{xcolor}
\usepackage{tikz}
\usepackage{algorithm}
\usepackage{algorithmic}
\definecolor{unitblue}{RGB}{52,101,164}
\definecolor{prodorange}{RGB}{230,126,34}

\icmltitlerunning{MorphingNet}

\begin{document}

\twocolumn[
  \icmltitle{MorphingNet: Symbolic Regression Where the Network Is the Formula}

  \ifdefined\arxivbuild
  \begin{icmlauthorlist}
    \icmlauthor{Jincheng Gao}{hkustgz}
    \icmlauthor{Jiahe Yi}{hkustgz}
    \icmlauthor{Shuo Sun}{hkustgz}
  \end{icmlauthorlist}
  \icmlaffiliation{hkustgz}{Thrust of Financial Technology, The Hong Kong University of Science and Technology (Guangzhou), Guangzhou, China}
  \icmlcorrespondingauthor{Shuo Sun}{shuosun@hkust-gz.edu.cn}
  \else
  \begin{icmlauthorlist}
    \icmlauthor{Anonymous}{anon}
  \end{icmlauthorlist}
  \icmlaffiliation{anon}{Anonymous Institution}
  \icmlcorrespondingauthor{Anonymous}{anonymous@example.com}
  \fi
  \icmlkeywords{Symbolic Regression, Scientific Discovery, Interpretability}

  \vskip 0.3in
]

\printAffiliationsAndNotice{}

\begin{abstract}
  We introduce MorphingNet, a network for symbolic regression (SR) whose units change their function continuously.
  Instead of searching over expressions built from a fixed operator library, MorphingNet trains the form of each function as a parameter,
  so SR becomes one continuous optimization problem.
  Its units are either pole-free rational functions or complex exponential--logarithm functions,
  which move between exponentials, sinusoids, damped oscillations, power laws and logarithms as their parameters change.
  The network is a readable formula at every step of training, and no operator library is needed to train it;
  a structure-recovery step afterwards rewrites the trained formula into short elementary expressions.
  Training is dense linear algebra, so it is GPU-friendly and scales to large data.
  MorphingNet achieves state-of-the-art accuracy on SRBench, recovers most generating formulas on synthetic benchmarks,
  and on large data is up to two orders of magnitude faster than the strongest search method.
  By turning the search for a formula into the training of a network, MorphingNet offers a new paradigm for SR and for scientific discovery from data.
\end{abstract}

\section{Introduction}

Symbolic regression (SR) looks for a closed-form expression $y=f(\mathbf{x})$ that fits data.
It chooses both the structure of the expression and its constants.
Because the answer is a formula, people can read it, check it against theory and reuse it.
SR has recovered physical laws from measurements~\citep{schmidt2009distilling,udrescu2020aifeynman}
and extracted equations from trained networks in astrophysics~\citep{cranmer2020discovering}.
In materials science it has rediscovered the Johnson--Mehl--Avrami--Kolmogorov transformation law and a Landau free energy~\citep{wang2019materials}.
Most SR methods are genetic programming (GP)~\citep{koza1992gp}.
GP evolves a population of expression trees built from an operator library, such as $\{+,\times,\sin,\exp\}$.
Each generation mutates and recombines the trees and scores every one on the data.
This search is slow.
Its cost grows with the number of samples, and data sets of $10^5$ to $10^6$ samples are now common.
On a million samples, Operon~\citep{burlacu2020operon}, one of the fastest GP systems, takes a median of 11.8 hours per problem in our experiments.

A natural idea is to run SR on a GPU, as deep learning does.
This is harder than it sounds.
A GPU is fast when it repeats one operation on large, uniformly shaped arrays.
A GP population holds many trees of different shapes, and each tree is a different program.
Recent systems still move evaluation, constant fitting and even variation onto the GPU~\citep{zhang2022gpusr,devries2025kozax,wu2025evogp,mao2026cusr}.
These systems make each evaluation faster, but the search itself stays discrete.
Every step proposes new trees, and each must be evaluated on all the data.
A GPU speeds up the arithmetic, not the number of trees the search must try.
What a GPU does best is continuous optimization:
one model, updated by one dense step on all the data, repeated until convergence.
Deep learning scales because its training has this form, and GP does not.

So the search should become training.
A line of work does exactly this by making the network itself the formula.
EQL~\citep{martius2016eql,sahoo2018eql} and its successors~\citep{li2024dysymnet,prunesymnet2024}
build networks whose units are elementary operators and train them by gradient descent.
ParFam~\citep{scholl2025parfam} fits rational functions composed with preselected elementary functions.
These are elegant steps toward SR as continuous optimization.
They optimize continuously, but the functions available to their units are chosen in advance.
In scientific discovery, we may not know which functions to choose.
An unfamiliar relationship may have no simple expression in the chosen library, even when its shape is simple.
This motivates a different approach: learn the shape of each function first, then seek a familiar symbolic description.
The learned shape remains available for inspection even when no such description is found.

The Kolmogorov--Arnold Network (KAN)~\citep{liu2025kan,liu2024kan2} takes this route.
It places a learnable spline on every edge.
Splines are flexible, and their shape is optimized continuously, like any network weight.
The library is used only at the end, to replace each spline by a matching symbolic function.
But a spline gains flexibility only by adding knots, and every knot is a constant.
So accuracy and length grow together.
Written out, a KAN is a spline soup: hundreds of piecewise constants, nested layer by layer.
The symbolic matching that shortens it is approximate, so the exported formula loses accuracy.

This raises a question.
Can SR be a continuous optimization problem that runs at scale,
and still return a formula that people can read: a short expression, or one whose structure, such as an inner product, is visible?

Our answer is a network whose units change their function continuously.
We call it MorphingNet.
A unit's function is not picked from a list.
It morphs from one form into another, for example from an exponential into a damped sinusoid,
as training moves its parameters.
So the form of the function is learned like any weight, and the network stays a closed-form expression throughout.
Like KAN, MorphingNet learns function shapes from data,
but each learned unit is itself a closed-form expression that can be plotted, inspected and then simplified.
Such units need a basis as flexible as the B-spline, but one whose elements are closed-form functions.
We use two families.
The first is the rational function.
Rationals approximate far better than polynomials of the same size:
they reach $|x|$ at the root-exponential rate $e^{-c\sqrt{n}}$ where polynomials reach it at an algebraic one~\citep{newman1964rational},
and Pad\'e approximants represent analytic functions with few constants~\citep{baker1996pade}.
Rational activations have already been used inside neural networks~\citep{molina2020pau,aghaei2024rkan}.
The second family is the exponential of a complex logarithm.
One formula with a few continuous parameters covers exponentials, sinusoids, damped oscillations and power laws.
Its closest relative is the soft exponential activation~\citep{godfrey2016softexp},
which interpolates between logarithm, identity and exponential and becomes sinusoidal when its parameter is complex.
Our unit composes its logarithmic and exponential branches, so that power laws and damped oscillations appear in the same unit.
Unlike a spline, each of these functions is global: a few parameters describe it on the whole domain.
Each unit first projects its inputs onto a learned direction, as in projection pursuit~\citep{friedman1981ppr},
and then applies a rational or complex exponential--logarithm function.
Product ports let units multiply as well as add.
The whole network is a smooth least-squares problem,
so we train it with a second-order method and prune it with a sparse group penalty.
Every training step is a dense matrix operation, so it runs on CPU and GPU alike.
The trained network is already a real-valued formula and exports without approximation.

As trained, these formulas are accurate but long.
No library is used during training, so a rational unit cannot write $\sin$ directly.
To our surprise, the units learn approximation schemes of their own.
A rational unit represents a sinusoid or an exponential by a Pad\'e-type rational function,
and redundant branches cancel each other.
These long formulas hide short ones.
This motivates a structure-recovery step that rewrites the trained formula under a fixed error bound.
It replaces approximations by elementary functions and recovers inner products.
A library enters only here, and its role differs from its role in GP.
It does not search for the formula; the formula already exists and is accurate.
The library only rewrites it in a shorter vocabulary, and users can extend it for their domain.
In one case, a trained formula with 48 operations and 519 constants, fitted to a million samples in 50 dimensions,
became $0.796\,e^{z_2}\sin z_1$.
Here $z_1$ and $z_2$ are learned linear combinations of the inputs, and the 45 irrelevant inputs dropped out of them.

The results bear out all three requirements.
On the \textbf{130 ground-truth problems} of SRBench~\citep{lacava2021srbench}, over \textbf{ten seeds},
MorphingNet reaches $R^2>0.999$ on \textbf{96 to 97\%} of problems at the three lower noise levels,
against at most 88\% for the published methods, and on 70\% at the highest level, against at most 68\%.
On 104 targets whose functions lie in the library, structure recovery returns the generating formula in \textbf{96 cases},
with median 7 operations and 10 constants against 16 and 16 for Operon.
On a million samples, MorphingNet runs in about five minutes on a GPU, \textbf{123 times faster} than Operon.

Our contributions are:
\begin{enumerate}
  \item \textbf{Morphing units.} A network of projections followed by rational and complex exponential--logarithm units,
    joined by product ports. The form of each function is optimized continuously,
    and the trained network is a formula without loss (Section~\ref{sec:network}).
  \item \textbf{Accuracy.} The highest accuracy among the SRBench methods on its ground-truth problems, at every noise level,
    with ablations that trace the gain to projection, and product ports to multiplicative targets
    (Sections~\ref{sec:srbench} and~\ref{sec:ablation}).
  \item \textbf{Structure recovery.} A post-training rewrite that shortens formulas under a fixed error bound
    and recovers the generating formula in most cases.
    A library is used only here, and can be extended by the user (Sections~\ref{sec:recovery} and~\ref{sec:recovery-exp}).
  \item \textbf{Scale.} Second-order training on CPU and GPU that matches GP at small scale
    and is one to two orders of magnitude faster on large data (Sections~\ref{sec:implementation} and~\ref{sec:scaling}).
\end{enumerate}

\section{Related Work}
\label{sec:related}

\paragraph{Search over expressions.}
Genetic programming~\citep{koza1992gp} evolves expression trees built from an operator library.
It found free-form physical laws~\citep{schmidt2009distilling},
and modern systems include PySR~\citep{cranmer2023pysr}, Operon~\citep{burlacu2020operon}
and GP-GOMEA~\citep{virgolin2021gomea}; the last two are among the strongest methods on SRBench~\citep{lacava2021srbench}.
AI Feynman~\citep{udrescu2020aifeynman} adds physics-inspired decompositions before the search.
Neural networks guide the search~\citep{landajuela2022udsr}, pretrained transformers propose expressions directly~\citep{biggio2021nesymres,kamienny2022e2e},
and large language models propose them inside an evolutionary loop~\citep{shojaee2025llmsr}.
These methods return short formulas, and they return them in the vocabulary of the library.
A target outside the library can only be approximated.
The exact solution of the Colebrook equation for pipe friction needs the Lambert $W$ function~\citep{brkic2011colebrook},
and a GP search without $W$ in its library returned logarithmic approximations~\citep{praks2018colebrook}.
A transformer trained on recurrent sequences, asked for a Bessel function, returned its asymptotic form~\citep{dascoli2022recurrent}.
Benchmarks build the right library in: for its ground-truth track, SRBench adds to every method the operators these problems need~\citep{lacava2021srbench}.
Every candidate is a discrete object, so constants are fitted inside a combinatorial search,
and the cost of that search grows with the size of the data.
MorphingNet does not search over expressions.
It fits one smooth model, and a library appears only when the fitted formula is shortened.
Any library from this line of work can be used at that step.

\paragraph{Networks of fixed operators and parametric families.}
EQL~\citep{martius2016eql,sahoo2018eql} turned symbolic regression into network training.
Its units are elementary operators, it is trained with a sparsity penalty, and the pruned network is the formula.
DySymNet~\citep{li2024dysymnet} searches the layout of such networks with a learned controller,
and PruneSymNet~\citep{prunesymnet2024}, SymbolNet~\citep{tsoi2025symbolnet} and LIES~\citep{montazerin2025lies} prune them toward compact expressions.
ParFam~\citep{scholl2025parfam} fits rational functions composed with preselected elementary functions by global continuous optimization.
TIR~\citep{defranca2022tir} evolves models of a similar rational form.
MorphingNet keeps the pipeline these methods share: a network that is a formula, a sparsity penalty, pruning and refitting.
What it changes is the unit.
In all of these methods the elementary function of a unit is chosen from a list,
either fixed in advance or picked by an outer search.
The choice of function is still discrete, even when the weights are trained by gradients.
In MorphingNet the function itself has continuous parameters, so no list is needed during training.

\paragraph{Continuous function forms.}
KAN~\citep{liu2025kan} learns a spline on every edge, and KAN~2.0~\citep{liu2024kan2} adds multiplication nodes;
softly symbolified KANs mix symbolic and spline terms during training~\citep{bagrow2025softly}.
Rational activations~\citep{molina2020pau,aghaei2024rkan} and the soft exponential~\citep{godfrey2016softexp}
make the shape of an activation trainable inside an ordinary network.
GINN-LP~\citep{ranasinghe2024ginnlp} learns the exponents of products of powers continuously, a family limited to Laurent polynomials.
Symbolic metamodels~\citep{alaa2019metamodels} fit Meijer G-functions, a closed-form family with continuous parameters,
and Symbolic Pursuit~\citep{crabbe2020pursuit} fits sums of them over projected inputs, both to explain a trained black-box model.
These methods all optimize the form of a function continuously.
They differ in what they produce at the end.
A spline can be written out as a formula: a piecewise polynomial with one constant per knot.
With many knots across many edges, that formula is too long to read.
KAN therefore replaces each spline by a matching symbolic function, and this step loses accuracy.
Rational and soft-exponential activations were not designed to be exported as formulas.
Symbolic Pursuit adds one ridge function at a time and has no products or composition.
GINN-LP does return a formula, but only a Laurent polynomial in the raw inputs, with no projections.
MorphingNet combines the three ingredients these methods have separately:
closed-form units with continuous parameters, projections of the inputs, and products between units.
It trains them jointly as one formula.

\section{Method}
\label{sec:method}

\begin{figure*}[t]
  \centering
  \resizebox{\textwidth}{!}{\input{figs/method_tikz.tex}}
  \caption{MorphingNet at a glance.
  (a) A two-layer network. Each layer reads the whole bank through an affine map; its ports feed units (blue) and product ports (orange),
  and their outputs are appended to the bank. Dark lines show the part that survives training on $y=2\sin(x_1)\,x_2+1$;
  gray parts were set to zero.
  (b) The complex exponential--logarithm unit $J$ at five settings of $(\alpha,\omega,\beta)$, computed from its definition.
  One formula moves continuously between these shapes.}
  \label{fig:method}
\end{figure*}

We start with an example, highlighted in Figure~\ref{fig:method}a. Suppose the data come from $y = 2\sin(x_1)\,x_2 + 1$.
MorphingNet can represent this function in two layers.
In the first layer, one unit reads the direction $z = w x_1$ and computes a function close to $\sin z$.
In the second layer, a product port multiplies that unit by $x_2$.
The output head scales the product by 2 and adds 1.
Every operation is an ordinary real operation, so the trained network can be written down as this formula.

The method has four steps, summarized in Algorithm~\ref{alg:morphsr}.
(1) Build a network whose units are closed-form functions with continuous parameters (Section~\ref{sec:network}).
(2) Train it so that unneeded units become exactly zero, then prune and refit at several levels (Section~\ref{sec:training}).
(3) Choose one formula from the resulting candidates (Section~\ref{sec:selection}).
(4) Rewrite that formula into a shorter one with the same accuracy (Section~\ref{sec:recovery}).
We follow the example through all four steps.

\begin{algorithm}[t]
  \caption{MorphingNet. Hyperparameters and capacities are listed in Appendix~\ref{app:hyper}.}
  \label{alg:morphsr}
  \begin{algorithmic}[1]
    \FOR{capacity $c$ = 1, 2, 3 layers}
      \STATE Fit: proximal LM on the loss~\eqref{eq:loss}, penalty off for the warm-up \hfill \textit{find the structure}
      \FOR{each threshold pair $(t,u)$}
        \STATE Mask coefficients below $t$ and units below $u$; refit with LM \hfill \textit{one candidate per level}
      \ENDFOR
      \STATE Export and replay every candidate; record length and selection error
      \IF{some candidate so far has selection error $\le\tau$}
        \STATE \textbf{break} \hfill \textit{capacity ladder}
      \ENDIF
    \ENDFOR
    \STATE Choose the shortest candidate on the length--error frontier within $\max(1.1\,e_{\min},\tau)$
    \STATE Apply rewrites (Table~\ref{tab:rewrites}) while the error stays within $\max(1.1\,e_0,\tau)$
  \end{algorithmic}
\end{algorithm}

\subsection{The network}
\label{sec:network}

Figure~\ref{fig:method}a shows the full network; the dark path is the example above.
Inputs are scaled to $[-1,1]$ by their declared ranges, $\tilde x = (x-c)/h$.
The network keeps a \emph{bank}: the list of all quantities computed so far.
It starts as $b^{(0)} = \tilde x$.
Each layer reads the whole bank through one affine map, $z = A b^{(\ell-1)} + a$.
Its outputs are of two kinds.
A \emph{unit} takes one port $z_j$ and applies a closed-form function $h_j = \varphi_{\theta_j}(z_j)$ with its own parameters $\theta_j$.
A \emph{product port} takes two ports and multiplies them, $z_a z_b$.
The layer appends all unit outputs and products to the bank,
so a later layer can read the raw inputs as well as every earlier result.
The output is an affine function of the final bank, $\hat y = m + s\,(v^\top b^{(L)} + v_0)$,
where $m$ and $s$ undo the scaling of $y$.

Each piece has a reason.
Because each unit reads a learned direction rather than a single coordinate,
a function such as $\sin(x_1+x_2)$ needs one unit, not a composition of one-dimensional pieces.
Products give multiplication and inner products $(u^\top x)(v^\top x)$ in one step.
The bank lets a later layer reuse an input or an earlier result without computing it again,
so a formula like $\sin u + v w$ does not need a copy of $u$ in every layer.

\paragraph{Rational unit (R).}
The rational unit is
\begin{equation}
  \varphi(z) = \frac{P(z)}{1+Q(z)^2},
\end{equation}
where $P$ and $Q$ are polynomials of degree $D$.
A plain ratio $P/Q$ would be more flexible, but it has poles, and a formula with a pole can divide by zero on new data.
The denominator $1+Q^2$ is at least one, so the unit is defined and smooth on the whole real line.
It represents $1/(1+z^2)$ and polynomials exactly, and approximates other shapes in a Pad\'e-like way~\citep{baker1996pade}.
In the example, a rational unit cannot write $\sin z$; it fits a rational approximation of it,
which Section~\ref{sec:recovery} later replaces.

\paragraph{Complex exponential--logarithm unit (C).}
The second family is built from one complex atom.
Let $L_\beta(z) = \log(1+\beta z)/\beta$, with $L_0(z) = z$, and let $q = \alpha + i\omega$ be complex. The unit is
\begin{equation}
  J(z) = \frac{e^{\,q L_\beta(z)} - 1}{q},
\end{equation}
and both its real and imaginary parts enter the bank.
Its parameters move it continuously between familiar functions:
\begin{equation}
  J(z) =
  \begin{cases}
    (e^{\alpha z}-1)/\alpha, & \beta=0,\ \omega=0,\\
    \bigl(\sin\omega z + i(1-\cos\omega z)\bigr)/\omega, & \beta=0,\ \alpha=0,\\
    (e^{\alpha z}e^{i\omega z}-1)/q, & \beta=0,\\
    \bigl((1+\beta z)^{q/\beta}-1\bigr)/q, & \beta\neq 0,\\
    \log(1+\beta z)/\beta, & q\to 0.
  \end{cases}
\end{equation}
From top to bottom these are an exponential, a sinusoid, a damped oscillation, a power law and a logarithm.
The division by $q$ keeps the atom finite as $q\to 0$, so the transitions are smooth.
A library that contains these five functions represents the same shapes.
The difference is how one moves between them.
In a library, going from $\exp$ to a damped oscillation is a discrete swap.
Here it is a small change of $\omega$, so the form of the function is trained by the same steps as every weight.
In the example, a complex unit with $\alpha=\beta=0$ gives $\sin(\omega z)/\omega$ exactly.

A run trains three capacities, with one, two and three layers (Appendix~\ref{app:hyper}).
Units and product ports are deliberately more than needed; training removes the unused ones.

\subsection{Training}
\label{sec:training}

Training has to do two things at once.
It has to fit the data to high precision, because recovering an exact formula needs errors far below those usual in neural network training.
And it has to make unneeded units exactly zero, so that they can be deleted from the formula.
The direct approach is Adam with an $\ell_1$ penalty, as in EQL.
It meets neither goal well.
Subgradient steps leave many coefficients small but nonzero,
and first-order methods slow down sharply near the optimum of an ill-conditioned least-squares problem.
Our networks are small, with a hundred to two thousand parameters, so we can afford a second-order method instead.

The objective is
\begin{equation}
  \mathcal L(\theta) = \frac{1}{n}\sum_{i=1}^n \bigl(\hat y_i - y_i\bigr)^2
  + \lambda_1 \|\theta_S\|_1 + \lambda_g \sum_{j} \|\theta_{\mathrm{out}(j)}\|_2 ,
  \label{eq:loss}
\end{equation}
with $y$ in scaled units.
The $\ell_1$ term acts on affine weights and unit coefficients.
The group term acts on the outgoing weights of each unit $j$: every weight with which a later layer or the head reads that unit.
When the group reaches zero, nothing reads the unit, and it is removed.
In the example, the smallest capacity that fits has 6 units per layer and the formula needs one.
The group term drives the outgoing weights of the other five to zero.

We minimize~\eqref{eq:loss} with a proximal Levenberg--Marquardt (LM) method~\citep{levenberg1944,marquardt1963}.
We want a step that decreases the least-squares error quickly and lands exactly on zero where the penalty asks for it.
Linearizing the residual, $r(\theta+\delta)\approx r+J\delta$, gives the subproblem
\begin{equation}
  \min_\delta\; \tfrac12\,\delta^\top (J^\top J + \mu I)\,\delta + g^\top\delta + \Omega(\theta+\delta),
\end{equation}
where $g = J^\top r$, $\Omega$ is the penalty and $\mu$ is the damping.
The quadratic part is the usual LM model, and $\mu$ acts as a trust region:
it shrinks when the predicted decrease is matched and grows when a step is rejected.
The penalty part is not differentiable at zero, which is what makes exact zeros possible.
We solve the subproblem with ADMM~\citep{boyd2011admm}, whose proximal step returns exact zeros (Appendix~\ref{app:admm}).

A fit first runs 200 steps without the penalty.
If the penalty were on from the start, it could remove a unit before the data had shown that the unit is needed.
The fit stops when the control error reaches a declared tolerance $\tau$, when the proximal gradient vanishes,
when the control error stops improving, or after 2000 steps.
The control error is measured on a separate split, not on the training data.
Snapshots that change the error or the structure are saved as candidates.

\paragraph{Pruning and refitting.}
After the fit, many coefficients are exactly zero but some are small and nonzero.
In the example, the first-layer port may keep a small weight on $x_2$ (a gray edge in Figure~\ref{fig:method}a).
Whether such a weight is noise or structure differs from problem to problem, so no single pruning level is right for all.
We therefore prune at a grid of levels and refit each.
A coefficient threshold $t$ removes coefficients below $t$ times the largest magnitude in their row,
and a unit threshold $u$ removes units whose group norm is below $u$ times the largest.
Both thresholds are measured on the same fit, so larger thresholds give nested masks.
Each pair gives one refit, which trains the surviving coefficients with plain LM under a fixed mask.
The grid has 31 refits per capacity (Algorithm~\ref{alg:morphsr}); identical masks share one.
A single refit at a middle level instead leaves fewer problems at the tolerance and longer formulas (Section~\ref{sec:ablation}).

\subsection{Choosing the formula}
\label{sec:selection}

Every candidate is exported and replayed: its formula is evaluated again and must agree with the network.
Each candidate has a complexity and an error on a selection split, which is used for nothing else.
We keep the candidates that no other candidate beats on both, the length--error frontier.
On the frontier, every candidate with selection error at most $\max(1.1\,e_{\min}, \tau)$ counts as tied,
and the tied candidate with the fewest unit operations, then the fewest constants, is chosen.
The factor 1.1 treats small differences as noise.
The tolerance $\tau$ treats all errors below it as equally good on noise-free data,
so extra constants are not spent on accuracy that has no meaning.

Capacities are trained in order.
When the fit and refits of a capacity are finished and a candidate of that or a smaller capacity has selection error at most $\tau$,
larger capacities are not trained, and their candidates do not enter selection.
This capacity ladder only fires when a small network already fits to $\tau$.
In the example, a one-layer network cannot multiply a unit by $x_2$, so its error stays above $\tau$.
The two-layer network reaches $\tau$, and the three-layer network is never trained.
Within the two-layer network, the refit that removed the $x_2$ weight and the one that kept it are both below $\tau$.
They tie, and the shorter one is chosen.
On noisy data, where the error cannot reach $\tau$, all three capacities run.
For noisy data the tie factor is applied after removing an estimate of the noise variance from the selection error (Appendix~\ref{app:noise}).
The test split is read only after the choice is saved.

\subsection{Structure recovery}
\label{sec:recovery}

The chosen formula is accurate but long.
A unit that cannot represent a function exactly approximates it, several units may share one direction,
and an inner product may be spread over several product ports.
Structure recovery rewrites the formula into a shorter one with the same accuracy.
Its ingredients are standard: candidate replacements, a joint refit and an error check.
What differs from search-based SR is the starting point.
Recovery starts from a formula that already fits, so it only has to shorten it, never to find it.

Each rewrite is a transaction.
It proposes a replacement, refits all remaining constants jointly with LM on the training split,
and is accepted only if the selection error stays below $\max(1.1\,e_0, \tau)$, where $e_0$ is the error of the original formula,
and the formula does not grow.
The bound is set once from the original formula, so a sequence of rewrites cannot drift away from it.
The rewrites are listed in Table~\ref{tab:rewrites}; they are applied in rounds until none is accepted.

\begin{table}[t]
  \caption{Rewrites used by structure recovery. Each is refitted jointly and kept only under the error bound.}
  \label{tab:rewrites}
  \vskip 0.05in
  \centering
  \small
  \setlength{\tabcolsep}{4pt}
  \begin{tabular}{p{0.24\columnwidth}p{0.68\columnwidth}}
    \toprule
    Rewrite & Action \\
    \midrule
    Unit & Replace a unit by $A\,\phi(z)+\gamma z+d$ with $\phi$ from the library.
      A C unit with $\beta=0$ becomes $K e^{\alpha z}\cos\omega z + M e^{\alpha z}\sin\omega z - K$ exactly. \\
    Merge & Replace several terms that depend on the same direction by one function. \\
    Quadratic & Collect products into a quadratic form and rewrite it with the fewest products by eigendecomposition. \\
    Delete & Delete terms; absorb linear terms into the bias of a product factor. \\
    Swap & Swap one elementary function for another if the error halves, e.g.\ $\tanh\to\exp$. \\
    Template & If the gradient outer product has rank two, fit $g(z_1)+h(z_2)$ or $g(z_1)\,h(z_2)$. \\
    \bottomrule
  \end{tabular}
\end{table}

In the example with rational units, the Unit rewrite replaces the rational approximation by $A\sin(z)+\gamma z+d$;
after the joint refit $\gamma$ and $d$ are negligible, the Delete rewrite removes them, and the formula becomes $2\sin(x_1)\,x_2+1$.
Exact canonicalizations, such as merging an affine map into its only reader, run after every step.
The library is used only here.
It contains the usual elementary functions and a few compositions of them (Appendix~\ref{app:recovery}); a user can extend it.

\subsection{Cost}
\label{sec:implementation}

One LM step costs $O(np^2)$ for the Gram matrix $J^\top J$ and $O(p^3)$ for its factorization,
where $n$ is the number of training rows and $p$ the number of parameters.
Every operation is dense and has a fixed shape, so the cost is linear in $n$ and maps directly onto a GPU.
The fit and the refits are independent and run in parallel.
CPU and GPU runs select the same formulas on the same data.
Appendix~\ref{app:ozaki} describes both implementations, including an exact integer scheme
that computes the Gram matrix on GPUs with weak double-precision arithmetic~\citep{ozaki2012error}.

\section{Experiments}
\label{sec:experiments}

The experiments answer four questions, one for each contribution.
Does MorphingNet fit benchmark targets more accurately than published methods (Section~\ref{sec:srbench})?
Does the gain come from projections and products (Section~\ref{sec:ablation})?
Does structure recovery find the generating formula (Section~\ref{sec:recovery-exp})?
How does the cost scale with data (Section~\ref{sec:scaling})?
Section~\ref{sec:baselines} compares with the neural and parametric methods of Section~\ref{sec:related} on a common suite.

\subsection{SRBench ground-truth problems}
\label{sec:srbench}

We use the \textbf{130 ground-truth problems} of SRBench~\citep{lacava2021srbench}, 116 from the Feynman lectures and 14 from Strogatz,
at the \textbf{four noise levels} of the benchmark and \textbf{ten random splits}.
Each split has a training set of at most 10\,000 rows, and separate control, selection and test sets.
We report the fraction of problems with test $R^2>0.999$ against the noise-free target, the standard SRBench accuracy criterion.
Published results are those of the SRBench release.

Table~\ref{tab:srbench} shows the result.
At the three lower noise levels the rational network reaches $R^2>0.999$ on \textbf{96 to 97\%} of the problems,
against 85 to 88\% for the best published methods.
At the highest noise level the strong methods are within a few points of each other, between 66\% and 70\%.
The spread over ten splits is one to three points.
Every rational formula is finite on every test set.
About 3\% of the confluent formulas are not finite at some test points, which counts as a miss.

\begin{table}[t]
  \caption{Fraction of SRBench ground-truth problems with test $R^2>0.999$, by noise level.
    MorphingNet: mean $\pm$ standard deviation over ten splits. Other methods: published SRBench results; MRGP~\citep{arnaldo2014mrgp}, SBP-GP~\citep{virgolin2019sbpgp}, AI Feynman~2.0~\citep{udrescu2020aifeynman2}.
    Best in each column in bold, second underlined.}
  \label{tab:srbench}
  \vskip 0.05in
  \centering
  \small
  \setlength{\tabcolsep}{4pt}
  \begin{tabular}{lcccc}
    \toprule
    Method & 0 & 0.001 & 0.01 & 0.1 \\
    \midrule
    MorphingNet-R & \textbf{97.0}$\pm$0.9 & \textbf{96.8}$\pm$1.1 & \textbf{96.3}$\pm$1.0 & \textbf{70.1}$\pm$2.5 \\
    MorphingNet-C & \underline{95.3}$\pm$1.5 & \underline{94.8}$\pm$1.0 & \underline{92.8}$\pm$1.6 & 66.4$\pm$3.5 \\
    \midrule
    MRGP & 88 & 87 & 83 & 3 \\
    Operon & 85 & 86 & 85 & \underline{68} \\
    SBP-GP & 75 & 74 & 74 & 50 \\
    AI Feynman & 74 & 73 & 69 & 12 \\
    GP-GOMEA & 71 & 72 & 72 & 67 \\
    \bottomrule
  \end{tabular}
\end{table}

\subsection{Where the accuracy comes from}
\label{sec:ablation}

We change one ingredient at a time and keep the rest of the pipeline fixed.
The suite has eight target families with five instances each, with and without noise (Table~\ref{tab:ablation}).
Projection is the ingredient that matters most.
If each first-layer port may read only one input coordinate, fewer noise-free problems reach the tolerance:
16 instead of 24 of 40 with rational units, and 18 instead of 24 with confluent units.
The median test error rises by $5.2\times$ and $2.2\times$.
The formulas also grow, because the network must rebuild each projection from single coordinates.
The median rational formula goes from 36 to 67 unit operations,
and the selected capacity moves to the largest network on 31 of 40 problems instead of 11.
On noisy data the error is set by the noise, and the two versions differ little ($1.3\times$ and $1.1\times$).

Product ports matter only where the target multiplies terms.
Without them, about as many noise-free problems reach the tolerance (22 and 23 of 40),
and the median formula is shorter (22 against 36 operations for rational units, 31 against 60 for confluent units).
Both unit families can emulate a product, for example as $\tfrac14\big((u+v)^2-(u-v)^2\big)$ or as a sum of exponentials,
so most targets do not need a dedicated port.
The exceptions are targets built around a product.
On $\sin u + vw$, rational units without products reach the tolerance on 1 of 5 instances instead of 4,
and the median error rises by $62\times$.
On damped oscillations $e^{-a(u+1)}\sin(bv+c)$, confluent units reach it on 2 instead of 4,
and the formula more than doubles its constants (842 against 370).
We keep the ports, because such targets are common in science.
On the other targets the price is a longer trained formula, not a less accurate one.
The pruning grid matters on noise-free data.
With a single refit at a middle level, 16 instead of 24 rational and 13 instead of 24 confluent problems reach the tolerance,
the median error rises by $1.9\times$ and $2.3\times$,
and the formulas grow (56 against 36 operations for rational units, 160 against 60 for confluent units).
On noisy data the error barely changes ($1.2\times$), but the formulas are still longer.
The grid costs about three times the CPU of a single refit.

\begin{table}[t]
\caption{Ablations on the current engine, eight families $\times$ five instances per condition.
Reached: problems whose validation error is within $\tau=10^{-5}$. Ratio: median paired test-error ratio to the full model.
Ops: median unit operations of the selected formula.}
\label{tab:ablation}
\centering\small
\setlength{\tabcolsep}{2.6pt}
\begin{tabular}{llrrrrr}
\toprule
 & & \multicolumn{3}{c}{Noise-free} & \multicolumn{2}{c}{Noisy} \\
\cmidrule(lr){3-5}\cmidrule(lr){6-7}
Unit & Variant & Reached & Ratio & Ops & Ratio & Ops \\
\midrule
Rational & full & 24/40 & 1 & 36 & 1 & 18 \\
 & first layer on axes & 16/40 & 5.2 & 67 & 1.3 & 30 \\
 & no products & 22/40 & 0.72 & 22 & 0.86 & 10 \\
 & one prune level & 16/40 & 1.9 & 56 & 1.2 & 27 \\
\midrule
Confluent & full & 24/40 & 1 & 60 & 1 & 17 \\
 & first layer on axes & 18/40 & 2.2 & 73 & 1.1 & 23 \\
 & no products & 23/40 & 1.0 & 31 & 0.89 & 10 \\
 & one prune level & 13/40 & 2.3 & 160 & 1.2 & 30 \\
\bottomrule
\end{tabular}
\end{table}

\subsection{Structure recovery}
\label{sec:recovery-exp}

To measure recovery, we need targets whose generating formula is known and expressible in the library.
We use 48 synthetic data sets in four families
(response $(a+bu)/(1+cv^2)$, oscillation $e^{-a(u+1)}\sin(bv+c)$, reuse $\sin u + vw$,
special $\mathrm{erf}(u)+c\,J_0(v)$), where $u,v,w$ are fixed projections of a few of the $d\in\{5,20,50\}$ inputs,
and $n$ runs from $10^3$ to $10^6$ rows.
We apply recovery to the 138 formulas chosen by the CPU and GPU runs, with the original error bound and no access to the test set.

Recovery returned the generating formula in 96 of the 104 cases whose functions are in the library,
with test error around $10^{-16}$ (Table~\ref{tab:recovery}).
All 36 response formulas were recovered, including one with 92 unit operations and 1478 constants,
which became a formula with 6 operations and 9 constants.
The irrelevant inputs were removed in every recovered case.
For example, an oscillation formula with 48 operations and 519 constants on $10^6$ rows and 50 inputs became
\begin{align*}
  y &= 0.796\, e^{z_2}\sin z_1,\\
  z_1 &= 0.621 - 0.234 x_2 - 0.173 x_3 + 1.72 x_4,\\
  z_2 &= 0.0647 x_1 - 0.121 x_2 + 0.0423 x_3 .
\end{align*}
The eight misses outside the special family stop at a nearby form:
five oscillation formulas keep a saturating $\tanh$ in place of $\exp$, with test error $2\times10^{-7}$ to $2\times10^{-5}$.
The special family uses $\mathrm{erf}$ and $J_0$, which are not in the library.
There the recovered formulas approximate them, with 8 to 13 operations and test error $6\times10^{-8}$ to $7\times10^{-7}$; Operon reaches $10^{-4}$.

Across all 138 formulas, the median recovered formula has 7 unit operations and 10 constants, against 16 and 16 for Operon on the same data.
Paired by data set, the recovered formula has no more constants than Operon in 108 cases, no more operations in 127, and lower test error in 137.

\begin{table}[t]
  \caption{Structure recovery on 138 formulas. Exact: test nRMSE $<10^{-12}$.
    Medians of unit operations / constants, before and after recovery, and of Operon on the same data.
    Bold: shorter than Operon in both counts.}
  \label{tab:recovery}
  \vskip 0.05in
  \centering
  \small
  \setlength{\tabcolsep}{4pt}
  \begin{tabular}{lcccc}
    \toprule
    Family & Exact & Before & After & Operon \\
    \midrule
    Response & 36/36 & 26/175 & \textbf{7/9} & 16/18 \\
    Oscillation & 29/34 & 26/250 & \textbf{7/8} & 17/14 \\
    Reuse & 31/34 & 15/45 & \textbf{5/11} & 17/21 \\
    Special & 0/34 & 21/115 & 13/23 & 13/14 \\
    \midrule
    All & 96/138 & 23/107 & \textbf{7/10} & 16/16 \\
    \bottomrule
  \end{tabular}
\end{table}

\subsection{Cost and scale}
\label{sec:scaling}

On the same 48 data sets we compare wall time with Operon, run with four threads, and PySR.
MorphingNet uses 16 single-threaded jobs on CPU, or three GPUs with four jobs each.
Table~\ref{tab:scaling} gives paired ratios.
On CPU, MorphingNet matches Operon at $10^3$ rows and is three to five times faster at $10^4$ and $10^5$.
PySR is one to two orders of magnitude slower than both.
On GPU the advantage grows with the data: 25 times at $10^5$ rows and \textbf{123 times at $10^6$},
where Operon takes a median of \textbf{11.8 hours} and MorphingNet \textbf{4.5 minutes} for R and 6.3 for C.
At every size the median test error of MorphingNet is \textbf{11 to 22 times lower} than that of Operon.
On the 8\,000-row development problem I.12.2, the GPU pipeline takes 50 seconds for R and 42 for C.

\begin{table}[t]
  \caption{Cost on 48 data sets. Medians of per-data-set ratios Operon / MorphingNet; above one favors MorphingNet.
    CPU at $10^6$ rows has two pairs.}
  \label{tab:scaling}
  \vskip 0.05in
  \centering
  \small
  \setlength{\tabcolsep}{4pt}
  \begin{tabular}{lcccc}
    \toprule
    Rows & $10^3$ & $10^4$ & $10^5$ & $10^6$ \\
    \midrule
    Wall time, CPU & 1.0 & 3.3 & 4.9 & 11.4 \\
    Wall time, GPU & -- & 4.4 & 24.9 & 122.6 \\
    Test error & 22 & 11 & 16 & 18 \\
    \bottomrule
  \end{tabular}
\end{table}

\subsection{Comparison with neural and parametric methods}
\label{sec:baselines}

We ran the closest methods of Section~\ref{sec:related} on a common suite of eight problem families,
with the same splits and the same selection rule for every method.
Table~\ref{tab:baselines} gives median test errors.
On noise-free data MorphingNet is two to three orders of magnitude more accurate than ParFam, EQL, KAN and DySymNet.
KAN loses about a factor of three between its splines and its exported formulas.
Brush~\citep{aldeia2026brush} and Operon are the strongest search methods.
Paired by problem, MorphingNet is more accurate than Brush on five of the eight in both conditions,
and more accurate than Operon on all eight noise-free problems and five of the eight noisy ones.

\begin{table}[t]
  \caption{Median test nRMSE on the common suite, eight problems per condition, and median unit operations / constants on noise-free data.
    Lowest error in bold, second underlined.}
  \label{tab:baselines}
  \vskip 0.05in
  \centering
  \small
  \setlength{\tabcolsep}{3pt}
  \begin{tabular}{lccc}
    \toprule
    Method & Noise-free & Noisy & Ops / const. \\
    \midrule
    MorphingNet-R & \textbf{7.9e-6} & \textbf{4.0e-3} & 30/130 \\
    MorphingNet-C & \underline{1.3e-5} & 4.8e-3 & 34/130 \\
    \midrule
    Brush & 3.9e-5 & 6.5e-3 & 20/17 \\
    Operon & 9.7e-5 & 5.5e-3 & 13/16 \\
    PySR & 2.4e-3 & 9.1e-3 & 20/13 \\
    ParFam & 9.4e-4 & \underline{4.5e-3} & 10/62 \\
    EQL & 9.5e-3 & 2.4e-2 & 31/46 \\
    KAN 2.0 (formula) & 1.5e-2 & 1.7e-2 & 33/34 \\
    KAN 2.0 (spline) & 4.9e-3 & 8.2e-3 & -- \\
    DySymNet & 1.5e-2 & 2.5e-2 & 10/15 \\
    \bottomrule
  \end{tabular}
\end{table}

\section{Discussion}
\label{sec:discussion}

\paragraph{Why trained formulas are long, and why that is not the end.}
A network of flexible units approximates what it cannot represent and spreads one structure over several branches.
This is a property of continuous structure learning, not of MorphingNet alone:
a wider KAN has more redundancy, not less.
What matters is whether the short formula can be recovered.
Because the trained network is already exact, recovery starts from an accurate formula and can be held to its error,
which is not possible when export itself is an approximation.

\paragraph{The role of the library.}
MorphingNet needs no operator library to train or to export.
The library enters only during recovery, and it decides how the result is named, not whether a structure is found.
The special family shows both sides: without $\mathrm{erf}$ and $J_0$ the formulas are approximations,
but they are two to three orders of magnitude more accurate than a search method with its library.
A user who expects such functions can add them.

\ifdefined\arxivbuild
\paragraph{What a formula learned from data can claim.}
Outside the range of the data, a formula found by MorphingNet can move far from the generating function.
We do not see this as a failure.
A data-driven method learns from the data and reports a formula that is faithful to them, and a finite sample is matched by many functions.
If $f$ fits the training points $x_1,\dots,x_n$, so does $g_c(x)=f(x)+c\prod_{i=1}^n(x-x_i)$ for every $c$.
The same holds within a tolerance on the whole domain:
$f(x)+\varepsilon\,((x-m)/r)^N$ stays within $\varepsilon$ of $f$ for $|x-m|\le r$ and grows like $|x|^N$ outside.
Different behaviour outside the domain is therefore a necessary consequence of learning from finite data.
\fi

\paragraph{Limitations.}
We did not study heavy-tailed noise, because SRBench uses Gaussian noise; this is next.
At the highest noise level MorphingNet leads by two points, within the spread over splits;
the gap at lower noise does not carry over because the selection split cannot resolve candidates below the noise.
The domain of the C unit with $\beta\neq0$ is checked on sampled points, not certified on the whole input box.
Structure recovery is greedy, takes a median of 215 seconds per formula on one core, and is sensitive to the last digits of its inputs;
a different execution order can end at a nearby form.
The fixed-operator variant of the network was not ported to the accelerated implementation.

\section{Conclusion}

MorphingNet makes the form of each function a continuous parameter, so symbolic regression becomes one smooth least-squares problem.
The trained network is the formula, it is accurate on standard benchmarks,
and a recovery step brings most formulas back to their generating form.
Because training is dense linear algebra, it runs at the speed of conventional symbolic regression on a CPU
and one to two orders of magnitude faster on a GPU when the data are large.

\ifdefined\arxivbuild
We end with why we think symbolic regression is promising for scientific discovery.
Our conjecture is that understanding the world is itself a large reverse-engineering problem.
Newton's law of gravitation was read off observations and checked against other theories, and its contribution is beyond question.
Yet suppose a creator made the world, gravity included, by some program.
Would Newton's law be that program? We think it need not be.
This takes nothing away from the law: within its domain it is excellent, and it fits with the rest of physics.
Much of science consists of such laws, valid under certain conditions or in a certain range, and this is the normal state of knowledge.
We hold discovered formulas to the same standard:
they need not be the true formula, if one exists, but should be accurate within the domain of the data and consistent with established theory in the field.
Judging the second condition, and choosing among the candidates from different seeds, is the work of domain experts;
MorphingNet is meant to give them candidates, not to replace that judgement.
\fi

\section*{Impact Statement}

This paper presents work whose goal is to advance the field of machine learning.
Symbolic regression is used to build interpretable models in science and engineering.
Formulas found from data can look authoritative while being wrong outside the range of the data;
we report the domain of every formula and recommend that discovered formulas be checked by domain experts before use.

\bibliography{refs}

\begin{thebibliography}{50}
\providecommand{\natexlab}[1]{#1}
\providecommand{\url}[1]{\texttt{#1}}
\expandafter\ifx\csname urlstyle\endcsname\relax
  \providecommand{\doi}[1]{doi: #1}\else
  \providecommand{\doi}{doi: \begingroup \urlstyle{rm}\Url}\fi

\bibitem[Aghaei(2024)]{aghaei2024rkan}
Aghaei, A.~A.
\newblock {rKAN}: Rational kolmogorov-arnold networks.
\newblock \emph{arXiv preprint arXiv:2406.14495}, 2024.

\bibitem[Alaa \& van~der Schaar(2019)Alaa and van~der
  Schaar]{alaa2019metamodels}
Alaa, A.~M. and van~der Schaar, M.
\newblock Demystifying black-box models with symbolic metamodels.
\newblock In \emph{Advances in Neural Information Processing Systems}, 2019.

\bibitem[Aldeia et~al.(2026)Aldeia, Romano, de~Fran{\c{c}}a, Herman, and
  La~Cava]{aldeia2026brush}
Aldeia, G. S.~I., Romano, J.~D., de~Fran{\c{c}}a, F.~O., Herman, D.~S., and
  La~Cava, W.~G.
\newblock Towards symbolic regression for interpretable clinical decision
  scores.
\newblock \emph{Philosophical Transactions of the Royal Society A}, 2026.
\newblock \doi{10.1098/rsta.2024.0588}.

\bibitem[Arnaldo et~al.(2014)Arnaldo, Krawiec, and O'Reilly]{arnaldo2014mrgp}
Arnaldo, I., Krawiec, K., and O'Reilly, U.-M.
\newblock Multiple regression genetic programming.
\newblock In \emph{Proceedings of the Genetic and Evolutionary Computation
  Conference}, 2014.

\bibitem[Bagrow \& Bongard(2025)Bagrow and Bongard]{bagrow2025softly}
Bagrow, J. and Bongard, J.
\newblock Softly symbolifying {Kolmogorov-Arnold} networks.
\newblock \emph{arXiv preprint arXiv:2512.07875}, 2025.

\bibitem[Baker \& Graves-Morris(1996)Baker and Graves-Morris]{baker1996pade}
Baker, G.~A. and Graves-Morris, P.
\newblock \emph{Pad\'e Approximants}.
\newblock Cambridge University Press, 2 edition, 1996.

\bibitem[Biggio et~al.(2021)Biggio, Bendinelli, Neitz, Lucchi, and
  Parascandolo]{biggio2021nesymres}
Biggio, L., Bendinelli, T., Neitz, A., Lucchi, A., and Parascandolo, G.
\newblock Neural symbolic regression that scales.
\newblock In \emph{International Conference on Machine Learning}, 2021.

\bibitem[Boyd et~al.(2011)Boyd, Parikh, Chu, Peleato, and
  Eckstein]{boyd2011admm}
Boyd, S., Parikh, N., Chu, E., Peleato, B., and Eckstein, J.
\newblock Distributed optimization and statistical learning via the alternating
  direction method of multipliers.
\newblock \emph{Foundations and Trends in Machine Learning}, 3\penalty0
  (1):\penalty0 1--122, 2011.

\bibitem[Brki{\'c}(2011)]{brkic2011colebrook}
Brki{\'c}, D.
\newblock {W} solutions of the {CW} equation for flow friction.
\newblock \emph{Applied Mathematics Letters}, 24\penalty0 (8):\penalty0
  1379--1383, 2011.

\bibitem[Burlacu et~al.(2020)Burlacu, Kronberger, and
  Kommenda]{burlacu2020operon}
Burlacu, B., Kronberger, G., and Kommenda, M.
\newblock Operon {C++}: An efficient genetic programming framework for symbolic
  regression.
\newblock In \emph{Proceedings of the Genetic and Evolutionary Computation
  Conference Companion}, pp.\  1562--1570, 2020.

\bibitem[Crabb{\'e} et~al.(2020)Crabb{\'e}, Zhang, Zame, and van~der
  Schaar]{crabbe2020pursuit}
Crabb{\'e}, J., Zhang, Y., Zame, W., and van~der Schaar, M.
\newblock Learning outside the black-box: The pursuit of interpretable models.
\newblock In \emph{Advances in Neural Information Processing Systems}, 2020.

\bibitem[Cranmer(2023)]{cranmer2023pysr}
Cranmer, M.
\newblock Interpretable machine learning for science with {PySR} and
  {SymbolicRegression.jl}.
\newblock \emph{arXiv preprint arXiv:2305.01582}, 2023.

\bibitem[Cranmer et~al.(2020)Cranmer, Sanchez-Gonzalez, Battaglia, Xu, Cranmer,
  Spergel, and Ho]{cranmer2020discovering}
Cranmer, M., Sanchez-Gonzalez, A., Battaglia, P., Xu, R., Cranmer, K., Spergel,
  D., and Ho, S.
\newblock Discovering symbolic models from deep learning with inductive biases.
\newblock In \emph{Advances in Neural Information Processing Systems}, 2020.

\bibitem[d'Ascoli et~al.(2022)d'Ascoli, Kamienny, Lample, and
  Charton]{dascoli2022recurrent}
d'Ascoli, S., Kamienny, P.-A., Lample, G., and Charton, F.
\newblock Deep symbolic regression for recurrent sequences.
\newblock In \emph{International Conference on Machine Learning}, 2022.

\bibitem[de~Fran{\c{c}}a(2022)]{defranca2022tir}
de~Fran{\c{c}}a, F.~O.
\newblock Transformation-interaction-rational representation for symbolic
  regression.
\newblock In \emph{Proceedings of the Genetic and Evolutionary Computation
  Conference}, 2022.

\bibitem[de~Vries et~al.(2025)de~Vries, Keemink, and van
  Gerven]{devries2025kozax}
de~Vries, S., Keemink, S.~W., and van Gerven, M. A.~J.
\newblock Kozax: Flexible and scalable genetic programming in {JAX}.
\newblock In \emph{Proceedings of the Genetic and Evolutionary Computation
  Conference Companion}, pp.\  603--606, 2025.

\bibitem[Friedman \& Stuetzle(1981)Friedman and Stuetzle]{friedman1981ppr}
Friedman, J.~H. and Stuetzle, W.
\newblock Projection pursuit regression.
\newblock \emph{Journal of the American Statistical Association}, 76\penalty0
  (376):\penalty0 817--823, 1981.

\bibitem[Godfrey \& Gashler(2016)Godfrey and Gashler]{godfrey2016softexp}
Godfrey, L.~B. and Gashler, M.~S.
\newblock A continuum among logarithmic, linear, and exponential functions, and
  its potential to improve generalization in neural networks.
\newblock \emph{arXiv preprint arXiv:1602.01321}, 2016.

\bibitem[Kamienny et~al.(2022)Kamienny, d'Ascoli, Lample, and
  Charton]{kamienny2022e2e}
Kamienny, P.-A., d'Ascoli, S., Lample, G., and Charton, F.
\newblock End-to-end symbolic regression with transformers.
\newblock In \emph{Advances in Neural Information Processing Systems}, 2022.

\bibitem[Koza(1992)]{koza1992gp}
Koza, J.~R.
\newblock \emph{Genetic Programming: On the Programming of Computers by Means
  of Natural Selection}.
\newblock MIT Press, 1992.

\bibitem[La~Cava et~al.(2021)La~Cava, Orzechowski, Burlacu, de~Fran{\c{c}}a,
  Virgolin, Jin, Kommenda, and Moore]{lacava2021srbench}
La~Cava, W., Orzechowski, P., Burlacu, B., de~Fran{\c{c}}a, F.~O., Virgolin,
  M., Jin, Y., Kommenda, M., and Moore, J.~H.
\newblock Contemporary symbolic regression methods and their relative
  performance.
\newblock In \emph{NeurIPS Track on Datasets and Benchmarks}, 2021.

\bibitem[Landajuela et~al.(2022)Landajuela, Lee, Yang, Glatt, Santiago,
  Aravena, Mundhenk, Mulcahy, and Petersen]{landajuela2022udsr}
Landajuela, M., Lee, C.~S., Yang, J., Glatt, R., Santiago, C.~P., Aravena, I.,
  Mundhenk, T., Mulcahy, G., and Petersen, B.~K.
\newblock A unified framework for deep symbolic regression.
\newblock In \emph{Advances in Neural Information Processing Systems}, 2022.

\bibitem[Levenberg(1944)]{levenberg1944}
Levenberg, K.
\newblock A method for the solution of certain non-linear problems in least
  squares.
\newblock \emph{Quarterly of Applied Mathematics}, 2\penalty0 (2):\penalty0
  164--168, 1944.

\bibitem[Li et~al.(2024)Li, Li, Yu, Wu, Sun, Liu, Li, Wei, Deng, and
  Hao]{li2024dysymnet}
Li, W., Li, W., Yu, L., Wu, M., Sun, L., Liu, J., Li, Y., Wei, S., Deng, Y.,
  and Hao, M.
\newblock A neural-guided dynamic symbolic network for exploring mathematical
  expressions from data.
\newblock In \emph{Proceedings of the 41st International Conference on Machine
  Learning}, volume 235 of \emph{PMLR}, pp.\  28222--28242, 2024.

\bibitem[Liu et~al.(2024)Liu, Ma, Wang, Matusik, and Tegmark]{liu2024kan2}
Liu, Z., Ma, P., Wang, Y., Matusik, W., and Tegmark, M.
\newblock {KAN} 2.0: Kolmogorov-arnold networks meet science.
\newblock \emph{arXiv preprint arXiv:2408.10205}, 2024.

\bibitem[Liu et~al.(2025)Liu, Wang, Vaidya, Ruehle, Halverson,
  Solja{\v{c}}i{\'c}, Hou, and Tegmark]{liu2025kan}
Liu, Z., Wang, Y., Vaidya, S., Ruehle, F., Halverson, J., Solja{\v{c}}i{\'c},
  M., Hou, T.~Y., and Tegmark, M.
\newblock {KAN}: Kolmogorov--arnold networks.
\newblock In \emph{International Conference on Learning Representations}, 2025.

\bibitem[Mao et~al.(2026)Mao, Liu, Lu, Zhao, Jiang, and Chen]{mao2026cusr}
Mao, H., Liu, X.~T., Lu, S., Zhao, P., Jiang, W., and Chen, Y.
\newblock Efficient constant optimization for symbolic regression with
  {GPU}-accelerated tree-based genetic programming.
\newblock In \emph{IEEE High Performance Extreme Computing Conference (HPEC)},
  2026.

\bibitem[Marquardt(1963)]{marquardt1963}
Marquardt, D.~W.
\newblock An algorithm for least-squares estimation of nonlinear parameters.
\newblock \emph{Journal of the Society for Industrial and Applied Mathematics},
  11\penalty0 (2):\penalty0 431--441, 1963.

\bibitem[Martius \& Lampert(2016)Martius and Lampert]{martius2016eql}
Martius, G. and Lampert, C.~H.
\newblock Extrapolation and learning equations.
\newblock \emph{arXiv preprint arXiv:1610.02995}, 2016.

\bibitem[Molina et~al.(2020)Molina, Schramowski, and Kersting]{molina2020pau}
Molina, A., Schramowski, P., and Kersting, K.
\newblock Pad{\'e} activation units: End-to-end learning of flexible activation
  functions in deep networks.
\newblock In \emph{International Conference on Learning Representations}, 2020.

\bibitem[Montazerin et~al.(2025)Montazerin, Al~Aawar, Ortega, and
  Srivastava]{montazerin2025lies}
Montazerin, M., Al~Aawar, M., Ortega, A., and Srivastava, A.
\newblock Sparse interpretable deep learning with {LIES} networks for symbolic
  regression.
\newblock \emph{arXiv preprint arXiv:2506.08267}, 2025.

\bibitem[Newman(1964)]{newman1964rational}
Newman, D.~J.
\newblock Rational approximation to $|x|$.
\newblock \emph{Michigan Mathematical Journal}, 11:\penalty0 11--14, 1964.

\bibitem[Ootomo et~al.(2024)Ootomo, Ozaki, and Yokota]{ootomo2024dgemm}
Ootomo, H., Ozaki, K., and Yokota, R.
\newblock {DGEMM} on integer matrix multiplication unit.
\newblock \emph{The International Journal of High Performance Computing
  Applications}, 38\penalty0 (4):\penalty0 297--313, 2024.

\bibitem[Ozaki et~al.(2012)Ozaki, Ogita, Oishi, and Rump]{ozaki2012error}
Ozaki, K., Ogita, T., Oishi, S., and Rump, S.~M.
\newblock Error-free transformations of matrix multiplication by using fast
  routines of matrix multiplication and its applications.
\newblock \emph{Numerical Algorithms}, 59\penalty0 (1):\penalty0 95--118, 2012.

\bibitem[Praks \& Brki{\'c}(2018)Praks and Brki{\'c}]{praks2018colebrook}
Praks, P. and Brki{\'c}, D.
\newblock Symbolic regression-based genetic approximations of the {Colebrook}
  equation for flow friction.
\newblock \emph{Water}, 10\penalty0 (9):\penalty0 1175, 2018.

\bibitem[Ranasinghe et~al.(2024)Ranasinghe, Senanayake, Seneviratne,
  Premaratne, and Halgamuge]{ranasinghe2024ginnlp}
Ranasinghe, N., Senanayake, D., Seneviratne, S., Premaratne, M., and Halgamuge,
  S.
\newblock {GINN-LP}: A growing interpretable neural network for discovering
  multivariate {Laurent} polynomial equations.
\newblock In \emph{Proceedings of the AAAI Conference on Artificial
  Intelligence}, volume~38, 2024.

\bibitem[Sahoo et~al.(2018)Sahoo, Lampert, and Martius]{sahoo2018eql}
Sahoo, S.~S., Lampert, C.~H., and Martius, G.
\newblock Learning equations for extrapolation and control.
\newblock In \emph{International Conference on Machine Learning}, 2018.

\bibitem[Schmidt \& Lipson(2009)Schmidt and Lipson]{schmidt2009distilling}
Schmidt, M. and Lipson, H.
\newblock Distilling free-form natural laws from experimental data.
\newblock \emph{Science}, 324\penalty0 (5923):\penalty0 81--85, 2009.

\bibitem[Scholl et~al.(2025)Scholl, Bieker, Hauger, and
  Kutyniok]{scholl2025parfam}
Scholl, P., Bieker, K., Hauger, H., and Kutyniok, G.
\newblock {ParFam} -- (neural guided) symbolic regression via continuous global
  optimization.
\newblock In \emph{International Conference on Learning Representations}, 2025.

\bibitem[Shojaee et~al.(2025)Shojaee, Meidani, Gupta, Farimani, and
  Reddy]{shojaee2025llmsr}
Shojaee, P., Meidani, K., Gupta, S., Farimani, A.~B., and Reddy, C.~K.
\newblock {LLM-SR}: Scientific equation discovery via programming with large
  language models.
\newblock In \emph{International Conference on Learning Representations}, 2025.

\bibitem[Tsoi et~al.(2025)Tsoi, Loncar, Dasu, and Harris]{tsoi2025symbolnet}
Tsoi, H.~F., Loncar, V., Dasu, S., and Harris, P.
\newblock {SymbolNet}: Neural symbolic regression with adaptive dynamic pruning
  for compression.
\newblock \emph{Machine Learning: Science and Technology}, 6\penalty0
  (1):\penalty0 015021, 2025.

\bibitem[Uchino et~al.(2025)Uchino, Ozaki, and Imamura]{uchino2025ozaki}
Uchino, Y., Ozaki, K., and Imamura, T.
\newblock Performance enhancement of the {Ozaki} scheme on integer matrix
  multiplication unit.
\newblock \emph{The International Journal of High Performance Computing
  Applications}, 39\penalty0 (3):\penalty0 462--476, 2025.

\bibitem[Udrescu \& Tegmark(2020)Udrescu and Tegmark]{udrescu2020aifeynman}
Udrescu, S.-M. and Tegmark, M.
\newblock {AI Feynman}: A physics-inspired method for symbolic regression.
\newblock \emph{Science Advances}, 6\penalty0 (16):\penalty0 eaay2631, 2020.

\bibitem[Udrescu et~al.(2020)Udrescu, Tan, Feng, Neto, Wu, and
  Tegmark]{udrescu2020aifeynman2}
Udrescu, S.-M., Tan, A., Feng, J., Neto, O., Wu, T., and Tegmark, M.
\newblock {AI Feynman 2.0}: Pareto-optimal symbolic regression exploiting graph
  modularity.
\newblock In \emph{Advances in Neural Information Processing Systems}, 2020.

\bibitem[Virgolin et~al.(2019)Virgolin, Alderliesten, and
  Bosman]{virgolin2019sbpgp}
Virgolin, M., Alderliesten, T., and Bosman, P. A.~N.
\newblock Linear scaling with and within semantic backpropagation-based genetic
  programming for symbolic regression.
\newblock In \emph{Proceedings of the Genetic and Evolutionary Computation
  Conference}, 2019.

\bibitem[Virgolin et~al.(2021)Virgolin, Alderliesten, Witteveen, and
  Bosman]{virgolin2021gomea}
Virgolin, M., Alderliesten, T., Witteveen, C., and Bosman, P. A.~N.
\newblock Improving model-based genetic programming for symbolic regression of
  small expressions.
\newblock \emph{Evolutionary Computation}, 29\penalty0 (2):\penalty0 211--237,
  2021.

\bibitem[Wang et~al.(2019)Wang, Wagner, and Rondinelli]{wang2019materials}
Wang, Y., Wagner, N., and Rondinelli, J.~M.
\newblock Symbolic regression in materials science.
\newblock \emph{MRS Communications}, 9\penalty0 (3):\penalty0 793--805, 2019.

\bibitem[Wu et~al.(2024)Wu, Li, Yu, Li, Liu, Li, and Hao]{prunesymnet2024}
Wu, M., Li, W., Yu, L., Li, W., Liu, J., Li, Y., and Hao, M.
\newblock {PruneSymNet}: A symbolic neural network and pruning algorithm for
  symbolic regression.
\newblock \emph{arXiv preprint arXiv:2401.15103}, 2024.

\bibitem[Wu et~al.(2025)Wu, Wang, Sun, Li, and Cheng]{wu2025evogp}
Wu, Z., Wang, L., Sun, K., Li, Z., and Cheng, R.
\newblock Enabling population-level parallelism in tree-based genetic
  programming for {GPU} acceleration.
\newblock \emph{arXiv preprint arXiv:2501.17168}, 2025.

\bibitem[Zhang et~al.(2022)Zhang, Lensen, and Sun]{zhang2022gpusr}
Zhang, R., Lensen, A., and Sun, Y.
\newblock Speeding up genetic programming based symbolic regression using
  {GPUs}.
\newblock In \emph{Pacific Rim International Conference on Artificial
  Intelligence (PRICAI)}, 2022.

\end{thebibliography}
\bibliographystyle{icml2026}

\appendix
\onecolumn

\section{Details of the units}
\label{app:units}

\paragraph{Rational unit.}
Each unit has numerator and denominator coefficients $p_0,\dots,p_D$ and $q_0,\dots,q_D$ and computes $P(z)/(1+Q(z)^2)$ by Horner's rule.
The denominator is at least one for every real $z$, so the unit needs no domain check.
At initialization $p_k\sim\mathcal N(0,0.025^2)$ with $p_1$ shifted by one, so each unit starts close to the identity,
and $q_k\sim\mathcal N(0,0.08^2)$.
The denominator coefficients must not start at zero, because the derivative of $1+Q^2$ with respect to $Q$ vanishes at $Q=0$.

\paragraph{One formula, no branches.}
The C unit is written with two functions that are analytic at the origin,
\begin{equation*}
  G(u)=\frac{\log(1+u)}{u},\qquad E(t)=\frac{e^{t}-1}{t},\qquad G(0)=E(0)=1,
\end{equation*}
as $L_\beta(z)=z\,G(\beta z)$ and $J(z)=L\,E(qL)$.
This is the definition of Section~\ref{sec:network} with the two divisions moved inside $G$ and $E$.
Both limits of that section follow at once: $\beta\to0$ gives $L\to z$, and $q\to0$ gives $J\to L$.
So the five cases listed there are values of one smooth function of $(z,q,\beta)$, and training moves through them without a switch.

\paragraph{Stable evaluation.}
Near the origin the quotients $G$ and $E$ lose all accuracy to cancellation.
We evaluate them by their Taylor series of degree 16 when $|u|\le 1/16$ or $|t|\le 1/2$, and otherwise by $\mathrm{log1p}(u)/u$ and $\mathrm{expm1}(t)/t$;
both branches are accurate to double rounding.
The derivatives used by LM are analytic:
\begin{equation*}
  \frac{\partial J}{\partial z}=\frac{e^{qL}}{1+\beta z},\qquad
  \frac{\partial J}{\partial q}=L^2E'(qL),\qquad
  \frac{\partial J}{\partial \beta}=e^{qL}\,z^2G'(\beta z),
\end{equation*}
with $E'$ and $G'$ evaluated by the same two-branch rule.
At initialization $\alpha\sim\mathcal N(0,0.2^2)$, $|\omega|=e^{s}$ with $s$ uniform on $[-1.4,1.4]$ and a random sign, and $\beta\sim\mathcal N(0,0.025^2)$.

\paragraph{Export.}
The exported formula is real.
With $\ell=L_\beta(z)$, $A=\mathrm{expm1}(\alpha\ell)\cos\omega\ell-2\sin^2(\omega\ell/2)$ and $B=e^{\alpha\ell}\sin\omega\ell$,
which are the real and imaginary parts of $e^{q\ell}-1$,
\begin{equation*}
  \operatorname{Re}J=\frac{\alpha A+\omega B}{\alpha^2+\omega^2},\qquad
  \operatorname{Im}J=\frac{\alpha B-\omega A}{\alpha^2+\omega^2}.
\end{equation*}
A unit with $\beta=0$ exports $\ell=z$, and a unit with $q=0$ exports $J=\ell$; these are the exact limits, not small substitutes.

\paragraph{Domain.}
For $\beta\neq0$ the logarithm needs $1+\beta z>0$.
Every accepted step is checked on the training rows and on 256 probes in the declared input box,
uniform points of which each has one coordinate set to its lower or upper bound.
A step is rejected if $1+\beta z\le0$ or $\alpha\,\ell>50$ at any of these points.
This guard is sampled, not a certificate over the box (Section~\ref{sec:discussion}).
Rational units need no guard.

\section{Hyperparameters}
\label{app:hyper}
Table~\ref{tab:capacity} lists the three capacities and Table~\ref{tab:hyper} the settings.
They are the same for every problem in the paper.
Affine weights start from $\mathcal N(0,0.04/k)$ for a layer that reads $k$ quantities, and biases from $\mathcal N(0,0.03^2)$.
Each run uses one initialization seed; the ten SRBench repetitions differ in their data splits.

\begin{table}[h]
  \caption{Capacities trained in each run, from small to large. Units and products are counted per layer.}
  \label{tab:capacity}
  \vskip 0.05in
  \centering
  \small
  \setlength{\tabcolsep}{4pt}
  \begin{tabular}{cccc}
    \toprule
    Layers & Units & Products & Degree $D$ \\
    \midrule
    1 & 6 & 2 & 3 \\
    2 & 6 & 2 & 3 \\
    3 & 12 & 4 & 5 \\
    \bottomrule
  \end{tabular}
\end{table}

\begin{table}[h]
  \caption{Training and selection settings.}
  \label{tab:hyper}
  \vskip 0.05in
  \centering
  \small
  \begin{tabular}{ll}
    \toprule
    Setting & Value \\
    \midrule
    $\ell_1$ weight $\lambda_1$ / group weight $\lambda_g$ & $3\times10^{-6}$ / $10^{-3}$ \\
    Warm-up without penalty & 200 LM steps \\
    Steps per fit and per refit & at most 2000 \\
    Control error & checked after every step \\
    Stagnation & 40 steps without a relative improvement of $10^{-5}$ \\
    Proximal gradient tolerance & $10^{-8}$ \\
    ADMM & residual tolerance $10^{-7}$, at most 400 inner iterations \\
    Tolerance $\tau$ (control and selection nRMSE) & $10^{-5}$ \\
    Domain probes & 256 \\
    Coefficient thresholds $t$ & $0, 0.05, 0.2, 0.35, 0.5, 0.65, 0.8, 0.9$ \\
    Unit thresholds $u$ & $0, 0.35, 0.65, 0.9$ \\
    Selection tie factor & 1.1 (on excess error for noisy data, Appendix~\ref{app:noise}) \\
    \bottomrule
  \end{tabular}
\end{table}

\section{The proximal step}
\label{app:admm}
The LM subproblem has a quadratic part and a nonsmooth penalty part. We split the two with ADMM~\citep{boyd2011admm}.
The quadratic part needs one Cholesky factorization of $J^\top J + (\mu+\rho) I$ per LM step.
The penalty part is a proximal map, a soft threshold followed by a group shrinkage, and it returns exact zeros.
The ADMM penalty $\rho$ is set by residual balancing.
Consecutive subproblems differ little, so we carry the dual variable and $\rho$ from one LM step to the next.
With this warm start the inner solver needs about ten iterations per step,
and it costs less than a tenth of the Jacobian and Gram computation.

\section{Selection on noisy data}
\label{app:noise}
On noisy data no candidate can have a selection error below the noise level $\sigma$, measured in the same scaled units.
The rule of Section~\ref{sec:selection} then loses its meaning.
Take $\sigma=0.1$ and a best candidate with $e_{\min}=0.1005$, whose error on the signal is $\sqrt{e_{\min}^2-\sigma^2}=0.01$.
The tie band $1.1\,e_{\min}$ admits a candidate whose error on the signal is $0.046$, five times larger.
We therefore apply the factor to the error that remains after the noise is removed.

The noise is estimated from the training split alone.
Inputs are standardized; for each row $i$ we fit a linear function to its $k=\max(4(d+1),20)$ nearest neighbours, excluding the row itself,
and take the residual $r_i$ of $y_i$ at $x_i$.
The local fit has error of its own, so the mean squared residual is divided by one plus the average leverage $(d+1)/k$ of the fit:
\begin{equation*}
  \hat\sigma^2=\frac{1}{1+(d+1)/k}\cdot\frac1n\sum_i r_i^2 ,
\end{equation*}
divided by the output scale.
With $s=\min(\hat\sigma,e_{\min})$, a frontier candidate with error $e$ is tied with the best if
\begin{equation*}
  e^2-s^2\le 1.21\,(e_{\min}^2-s^2),
\end{equation*}
and the shortest tied candidate is chosen as before.
When $\hat\sigma\ge e_{\min}$ the band closes and the most accurate candidate is chosen.
A local linear fit overestimates the noise where the target curves strongly within a neighbourhood.
So when $\hat\sigma>1.5\,e_{\min}$ the estimate is not trusted and the rule of Section~\ref{sec:selection} is used.
On SRBench this fallback fires on most problems at noise 0.001, where curvature dominates the local residual, and on fewer at higher noise.

The rule matters at the highest noise level.
There the plain rule reaches $R^2>0.999$ on 33\% of the problems with rational units and 23\% with confluent units, against 70\% and 66\% with this rule.
At noise 0.01 and below the two rules agree within one point (Table~\ref{tab:srbench-full}).

\section{Structure recovery}
\label{app:recovery}

\paragraph{Library.}
A one-dimensional piece $g(z)$ of the formula is replaced by $A\,\phi(z)+\gamma z+d$, with $\phi$ one of
$\sin(az+b)$, $e^{az}$, $(z+a)^2$, $1/(z+a)$, $\log(z+a)$, $\sqrt{z+a}$, $\tanh(az+b)$, $e^{-(az+b)^2}$, $1/(1+(az+b)^2)$, $(z+a)^b$ and $e^{az}\sin(bz+c)$,
or by a constant or a linear function.
For each form the linear coefficients $A,\gamma,d$ are solved by least squares inside a nonlinear least-squares fit of the inner parameters, from several starts.
Candidates are tried in a fixed order of preference, first constant and linear, then the other forms, with $(z+a)^b$ and the damped sine last, and within each group by fit error.
A user adds a form by giving its expression and its starting points.

\paragraph{Acceptance.}
A rewrite removes the parts of the formula it leaves unused and is accepted if the formula stays finite,
has no more operations and no more constants than before, with at least one of the two fewer,
and has selection nRMSE at most $\max(1.1\,e_0,\tau)$.
Every candidate is first checked as proposed.
If none passes, the first three are refitted jointly with all other constants (LM, 4000 rows, 50 iterations) and checked again.
Refits that do not change the structure are kept only if they lower the selection error.

\paragraph{Rewrites.}
\emph{Unit} replaces one unit. A C unit with $\beta=0$ becomes the exact real form of Appendix~\ref{app:units}.
\emph{Merge} collects all terms of one consumer that depend on the same input direction, such as several rational units that together approximate one sinusoid, into one function of that direction.
\emph{Quadratic} collects the products of affine ports read by one consumer into a symmetric matrix $Q$, a linear part and a constant.
Its eigendecomposition gives the fewest products: a positive and a negative eigenvector $e_+$, $e_-$ share one product $(e_++e_-)^\top x\,(e_+-e_-)^\top x$,
so $Q$ needs as many products as the larger of its numbers of positive and negative eigenvalues.
Small eigenvalues are dropped at relative thresholds from $10^{-9}$ to $10^{-2}$, each tried in turn.
If the consumer also holds a function of a direction $u$ in the range of $Q=V\Lambda V^\top$,
then removing $\kappa\,uu^\top$ with $\kappa=1/(c^\top\Lambda^{-1}c)$, $c=V^\top u$, lowers the rank of $Q$ by one.
The square $\kappa(u^\top x)^2$ is kept as a function of $u$, where Merge can combine it with the other terms on $u$.
\emph{Delete} removes affine terms one at a time with a local refit,
and absorbs a linear term $w\,(p^\top x)$ into the bias of a product factor when the shift reproduces it.
\emph{Swap} replaces one elementary function by another when the jointly refitted and pruned formula halves the selection error and is not longer.
\emph{Template} applies when the averaged gradient outer product $\mathbb E[\nabla f\,\nabla f^\top]$ on 4000 rows has rank two.
In that plane, the Hessians of an additive $g_1(a_1^\top x)+g_2(a_2^\top x)$ lie in the span of $a_1a_1^\top$ and $a_2a_2^\top$,
and so do those of $\log|f|$ for a product $g_1g_2$.
This fixes $a_1$ and $a_2$; each $g$ is chosen from the linear function and the first nine forms of the library,
and the template is kept if, refitted and pruned, it is within the bound and shorter than the current formula.

\paragraph{Order and cost.}
A round applies Unit, Merge, Quadratic, absorption, a joint refit, Delete, another refit and Swap, and then Template if the formula has a length not seen before.
Rounds repeat until none changes the formula, at most four times.
A last pass applies Swap to forms accepted in the final round, and biases below $10^{-12}$ of the output scale, left over by exact rewrites, are set to zero.
Recovery takes a median of 215 seconds per formula on one CPU core.

\section{Implementation}
\label{app:ozaki}
On CPU, a C++ core computes the analytic Jacobian for all rows at once,
and the fit and refits, which are independent, execute in parallel as single-threaded processes.
Sixteen single-threaded jobs use five times less CPU time than four jobs with four BLAS threads each, at a similar wall time.
On GPU, one fused kernel computes the forward pass and the Jacobian, and the data, Gram matrix, factorization and ADMM stay on the device.
Several fits share one GPU through MPS, which lets their kernels run concurrently.

\paragraph{Gram matrix on integer tensor cores.}
Consumer GPUs have weak double-precision arithmetic: one $8000\times1745$ Gram matrix takes 30\,ms in FP64 on an RTX~5090.
We compute it with the Ozaki scheme~\citep{ozaki2012error}, as on integer matrix units in~\citet{ootomo2024dgemm,uchino2025ozaki}.
Each row of $J^\top$, one parameter over all data rows, is scaled by a power of two so that its largest entry lies in $[64,128)$,
and split into eight slices of seven bits, stored as 8-bit integers.
The products of slices $t$ and $u$ with the same weight $t+u$ are summed on INT8 tensor cores with 32-bit integer accumulation.
The data rows are processed in blocks of 16\,384, since $8\cdot16\,384\cdot127^2<2^{31}$, so every accumulation is exact.
Symmetry halves the work, $G=H+H^\top+D$ with $H$ the pairs $t<u$ and $D$ the pairs $t=u$,
and weights $t+u\ge8$, which fall below double rounding, are dropped.
The levels are combined in double precision.
On Jacobians from real fits, the relative spectral-norm error is $4\times10^{-17}$ to $9\times10^{-15}$,
against $2\times10^{-16}$ to $2\times10^{-15}$ for an FP64 matrix product and about $10^{-7}$ in FP32.
This is $10^{-5}$ to $0.05$ times the smallest LM damping used, so it does not change the steps:
a 300-step refit reaches the same best control error as FP64 to seven digits.
The scheme is 8 to 10 times faster than FP64: 3.7 against 30\,ms for $8000\times1745$, and 28 against 241\,ms for 65\,536 rows.
An FP32 Gram matrix is faster still, but its error exceeds the damping by five orders of magnitude, and two thirds of the refit steps had to be repeated in FP64.

\section{Full results}
\label{app:results}
Table~\ref{tab:srbench-full} breaks down the SRBench result by problem set and gives the result of the plain selection rule.

\begin{table}[h]
  \caption{SRBench ground-truth problems: percentage with test $R^2>0.999$, mean $\pm$ standard deviation over ten splits,
  for all 130 problems, the 116 Feynman and the 14 Strogatz problems, and with the plain $1.1\,e_{\min}$ rule instead of the noise-aware rule.
  Last column: median unit operations / constants of the chosen formulas, before structure recovery.}
  \label{tab:srbench-full}
  \vskip 0.05in
  \centering
  \small
  \begin{tabular}{llccccc}
    \toprule
    Unit & Noise & All & Feynman & Strogatz & Plain rule & Ops / const. \\
    \midrule
    Rational & 0     & 97.0$\pm$0.9 & 97.8$\pm$0.4 & 90.0$\pm$8.0 & 97.0$\pm$0.9 & 59 / 216 \\
             & 0.001 & 96.8$\pm$1.1 & 97.8$\pm$0.6 & 88.6$\pm$9.1 & 96.8$\pm$1.1 & 39 / 94 \\
             & 0.01  & 96.3$\pm$1.0 & 98.0$\pm$0.4 & 82.1$\pm$8.0 & 96.3$\pm$1.0 & 36 / 83 \\
             & 0.1   & 70.1$\pm$2.5 & 76.5$\pm$2.9 & 17.1$\pm$5.7 & 33.2$\pm$3.5 & 30 / 61 \\
    \midrule
    Confluent & 0     & 95.3$\pm$1.5 & 96.5$\pm$1.0 & 85.7$\pm$9.6 & 95.3$\pm$1.5 & 50 / 196 \\
              & 0.001 & 94.8$\pm$1.0 & 95.7$\pm$0.8 & 87.9$\pm$6.4 & 94.8$\pm$1.0 & 37 / 100 \\
              & 0.01  & 92.8$\pm$1.6 & 94.2$\pm$1.4 & 81.4$\pm$5.7 & 92.2$\pm$2.4 & 35 / 92 \\
              & 0.1   & 66.4$\pm$3.5 & 71.2$\pm$4.0 & 26.4$\pm$8.5 & 22.8$\pm$2.4 & 27 / 60 \\
    \bottomrule
  \end{tabular}
\end{table}

\end{document}